\documentclass[conference]{IEEEtran}
\IEEEoverridecommandlockouts
\usepackage{cite}
\usepackage{amsmath,amssymb,amsfonts}
\usepackage{algorithmic}
\usepackage{graphicx}
\usepackage{textcomp}
\usepackage{xcolor}
\def\BibTeX{{\rm B\kern-.05em{\sc i\kern-.025em b}\kern-.08em
    T\kern-.1667em\lower.7ex\hbox{E}\kern-.125emX}}

\usepackage{tikz}
\usepackage{multirow}
\usepackage{multicol}
\usepackage{booktabs}
\usepackage{subcaption}

\input{Definitions.tex}
\begin{document}

\title{Self-supervised Learning for Anomaly Detection in Computational Workflows}

\author{\IEEEauthorblockN{Hongwei Jin, Krishnan Raghavan}
  \IEEEauthorblockA{\textit{Mathematics and Computer Science Division} \\
    \textit{Argonne National Laboratory}\\
    Lemont, IL USA \\
    \{jinh, kraghavan\}@anl.gov}
  \and
  \IEEEauthorblockN{George Papadimitriou}
  \IEEEauthorblockA{\textit{Information Sciences Institute} \\
    \textit{University of Southern California}\\
    Los Angeles, CA USA \\
    georgpap@isi.edu}
  \and
  \IEEEauthorblockN{Cong Wang}
  \IEEEauthorblockA{\textit{RENCI} \\
    \textit{University of North Carolina}\\
    Chapel Hill, NC USA \\
    cwang@renci.org}
  \and
  \IEEEauthorblockN{Anirban Mandal}
  \IEEEauthorblockA{\textit{RENCI} \\
    \textit{University of North Carolina}\\
    Chapel Hill, NC USA \\
    anirban@renci.org}
  \and
  \IEEEauthorblockN{Ewa Deelman}
  \IEEEauthorblockA{\textit{Information Sciences Institute} \\
    \textit{University of Southern California}\\
    Los Angeles, CA USA \\
    deelman@isi.edu}
  \and
  \IEEEauthorblockN{Prasanna Balaprakash}
  \IEEEauthorblockA{\textit{Computing and Computational Sciences Directorate} \\
    \textit{Oak Ridge National Laboratory}\\
    Oak Ridge, TN USA \\
    pbalapra@ornl.gov}
}

\maketitle

\begin{abstract}
  Anomaly detection is the task of identifying abnormal behavior of a system. Anomaly detection in computational workflows is of special interest because of its wide implications in various domains such as cybersecurity, finance, and social networks. However, anomaly detection in computational workflows~(often modeled as graphs) is a relatively unexplored problem and poses distinct challenges. For instance, when anomaly detection is performed on graph data, the complex interdependency of nodes and edges, the heterogeneity of node attributes, and edge types must be accounted for. Although the use of graph neural networks can help capture complex inter-dependencies, the scarcity of labeled anomalous examples from workflow executions is still a significant challenge. To address this problem, we introduce an autoencoder-driven self-supervised learning~(SSL) approach that learns a summary statistic from unlabeled workflow data and estimates the normal behavior of the computational workflow in the latent space. In this approach, we combine generative and contrastive learning objectives to detect outliers in the summary statistics. We demonstrate that by estimating the distribution of normal behavior in the latent space, we can outperform state-of-the-art anomaly detection methods on our benchmark datasets.
\end{abstract}

\begin{IEEEkeywords}
  computational workflow, anomaly detection, self-supervised learning
\end{IEEEkeywords}

\section{Introduction}
% \note{prediction workflow is not a scientific computing workflow, we'd better revise the ``scientific computing workflow'' to ``computational workflow''}

Scientific computing is a branch of computational science that involves the development and use of mathematical and computational models and simulations to study complex phenomena in various fields, such as physics, chemistry, biology, and engineering. These computational models and large-scale analyses are often executed on distributed and high-performance computing resources through a series of steps. These steps constitute a computational workflow. To ensure the accurate execution of these workflows, workflow management systems have emerged that manage the execution and validation. However, abnormal or erroneous patterns in the infrastructure still exist and can be detrimental to the performance of the computational workflows and to their turnaround time. Detecting anomalous behavior is therefore an extremely important paradigm for detecting and diagnosing faults in workflows (e.g., to detect deviations from normal behavior) and will enable scientists and system operators to improve the reliability and efficiency of workflow executions.

Nevertheless, the literature in anomaly detection for scientific workflows is sparse~\cite{jin2022workflow}.
Despite preliminary efforts, several key challenges still exist:
(1) there is no template for determining the anomaly in the workflow settings,
(2) most of the time, the training data comprises several orders of magnitude more data points corresponding to the normal behavior than the anomaly behavior.
Although these challenges have been addressed in various capacities in the past~\cite{borghesi2019anomaly, kiran2020detecting, herath2019ramp}, all these methods rely on predefined labels. However, determining/generating labeled data is usually a time-consuming task. Therefore, the most important challenge in anomaly detection is the development of algorithms that will perform unsupervised anomaly detection without using labeled data, especially when the input data is defined by graphs.

The key requirement for unsupervised anomaly detection is to characterize outliers in the distribution of the data. However, the characterization of distributions on graphs is a relatively unexplored idea. In this work we take a new approach to characterize distributions on scientific workflow anomaly data ---we extract summary statistics from the graph data and characterize outliers over the distribution of the summary statistics. To accomplish this, we develop a self-supervised learning (SSL) approach  comprising two neural networks, where one network acts as an encoder and the other network acts as a decoder.

We consider workflows to be represented as directed acyclic graphs (DAGs). Typically these DAGs consist of structure information in the form of the adjacency matrix and feature information in the form of a feature matrix. To capture the structure information while extracting summary statistics, the encoder network comprises a graph neural network~\cite{bielak2022graph}. A graph neural network will take as input the feature and adjacency matrix information to extract an embedding in the latent space of the encoder. Then, the decoder network takes input samples from the latent space and reconstructs the input feature matrix. Since the structure of workflows does not change, we reconstruct  only the feature matrix as the output of the decoder. By introducing variations into the latent space and constraining the latent layer, we allow for capturing the nature of anomalies in the latent space and thus achieve better detection of anomalies through the reconstruction error.

Typically, for unsupervised anomaly detection, where the availability of samples corresponding to abnormal behavior is often limited because of factors such as sample collection difficulties and the high cost of running workflows with anomalies, the use of data augmentation techniques can be beneficial. Data augmentation techniques can be used to generate synthetic samples that are similar to the positive samples. Thanks to the autoencoder framework~\cite{kingma2019introduction}, this kind of data augmentation can be achieved by introducing variations into the latent space and constraining the latent layer.

We characterize the distribution in the latent layer through two distributions. First, we utilize the standard normal to constrain the latent space such that the latent space is described by using a continuous random variable and identifies a rich representation for the distribution. Alternatively, we utilize the Gumbel softmax trick to characterize the latent space using a continuous approximation of discrete random variables where anomalies can be represented by values closer to zero and normal by values closer to one. Such structure in the latent space is enforced by combining a contrastive and generative learning objective. We show that these constraints not only provide a better reconstruction of the feature matrix but also improve model performance under different metrics. In contrast, many traditional approaches result in an out-of-memory error or require excessive computing resources to be effective in a workflow environment.

The key contribution of this paper is the development of a dedicated approach for detecting anomalies in scientific workflows, which offers several advantages over readily available methods. The specific contributions are as follows:
\begin{itemize}
  \item We develop self-supervised learning for an anomaly detection approach designed for scientific workflows, modeled as DAGs.
  \item We introduce the data augmentation to retrieve pseudo labels of original data and apply contrastive learning to improve the performance of anomaly detection;
  \item We propose the learnable normal and Gumbel distribution to constrain the latent space and improve the reconstruction of the feature matrix.
  \item We enhance the scalability of the approach by 2-3 times compared to readily available methods with 2--5\% improvement in ROC-AUC scores.
\end{itemize}

\section{Related Works}
While a massive amount of research exists for datasets in the form of images and text, and even when the data is described by graphs, the literature is sparse when considering workflow data.

Traditional approaches for anomaly detection involve statistical methods that seek to identify outliers in a distribution. Methods such as clustering~\cite{sohn2023anomaly}, principal component analysis (PCA)~\cite{ma2023mppca}, and auto-regressive integrated moving average models~\cite{goldstein2023special} are commonly utilized to detect anomalies. Many times, statistical approaches are not effective in characterizing complex distributions in data that are represented by graphs.

More recently, machine learning (ML) methods have emerged as promising alternatives to statistical methods in their ability to characterize complex data distributions. Typically, these approaches involve learning a map between the anomaly class and the input data such that anomalies can be directly identified. Numerous ML methods have been developed for both tabular and graph data. Popular methods include (graph) neural networks~\cite{yuan2021higher}, decision trees~\cite{breiman2001random}, and support vector machines (SVMs)~\cite{scholkopf2001estimating}.
Especially, recent applications~\cite{ouyang2020unified, feng2023unsupervised, kim2022graph, jin2022workflow, jin2023graph} of graph neural networks (GNNs)  in anomaly detection show that the graph structure is beneficial because of its feature extraction between the inputs, with both supervised and unsupervised learning.

Another class of methods involves hybrid methods that utilize the probabilistic efficiency of statistical methods while leveraging the ability of ML methods to characterize complex distribution. This approach has been shown to provide improved accuracy and robustness in anomaly detection~\cite{kriegel2008angle, khalique2021voa}. Despite excellent detection accuracy, however, these classes more often than not rely on labeled data to increase efficiency. However, generating labels is extremely expensive and time-consuming. One way of handling this situation is to intelligently generate the requisite labels as part of the training procedure in an unsupervised manner. These classes of methods are known as self-supervised learning (SSL) methods.

SSL is a machine learning paradigm that aims to learn useful representations from unlabeled data by generating supervisory signals from the data itself. In contrast to supervised learning, where the model is trained on labeled data, self-supervised learning does not require any explicit supervision. This feature is especially useful in cases where labeled data is scarce or expensive to obtain, such as in large computational workflows.
SSL has been applied to various domains such as natural language processing (NLP)~\cite{devlin2018bert}, computer vision~\cite{atito2021sit}, and reinforcement learning~\cite{ze2023visual}. Notably, SSL has been shown to be effective in learning representations for powerful large language models~\cite{devlin2018bert}.

Despite impressive results in many scenarios, the typical unsupervised learning approaches do not fit and scale well in the domain of computational workflows. In contrast, by capturing the structural information from the graph level and feature information from the node level, our approach can better characterize the distribution of anomalies effectively in the latent space of the encoder. Moreover, in contrast with the prior approaches, our approach  is extremely scalable with respect to the size of workflows.

\section{Anomaly Detection Methodologies}
We use bold uppercase letters (e.g., $\Avec$), bold lowercase letters (\eg\ $\avec$), and regular lowercase letters (e.g., $a$) to denote matrices, vectors, and scalars, respectively. Let $\Gcal = \cbr{\Avec, \Xvec}$ be an input attributed scientific workflow graph, where $\Avec \in \RR^{n \times n}$ and $\Xvec \in \RR^{n \times d}$ denote the adjacency matrix and attributed~(feature) matrix of $\Gcal$, respectively. The scientific workflows are modeled as DAGs, where $\Avec_{ij} = 1$ if there is a directed edge from node $i$ to node $j$. Without further notification, we preprocess the workflows as undirected graphs, that is, $\Avec_{ij} = \Avec_{ji}$.

The key aim of anomaly detection is to mark whether the nodes in $\Gcal$ are anomalous or not. Since there are $n$ nodes in $\Gcal,$ we define an anomaly score vector for each node as $\tilde{\yvec} \in [0,1]^{n}$ such that an anomaly score can be between $0$ and $1$. Then the goal of the ML approach is to learn a function
\begin{align}
  \tilde{\yvec} = h(\Gcal).
\end{align}

The goal of the rest of the paper is to construct $g$. In this work, we will decompose $h$ into two functions $g$ and $f$ such that $g(\Gcal)$ as $h(f(\Gcal))$, where the function $f$ is an encoder that transforms the input graph into a latent space and the transformation $h$ reconstructs the input feature matrix and then generates the anomaly score on the input data. This structure is reminiscent of the encoder-decoder setup also known as autoencoder. An autoencoder is a type of neural network that can learn to compress and decompress data. Since we do not have any targets on the anomalies, we will utilize a self-supervised learning paradigm~\cite{kingma2019introduction} to generate a variational latent space. The basic idea here is to use the encoder to generate pseudo labels on the data and train the latent space such that the distinction between normal and anomalies can be detected in the latent space.

\subsection{Step 1: Self-supervised Learning as an Encoder}
\begin{figure*}
  \centering
  \includegraphics[width=.75\linewidth]{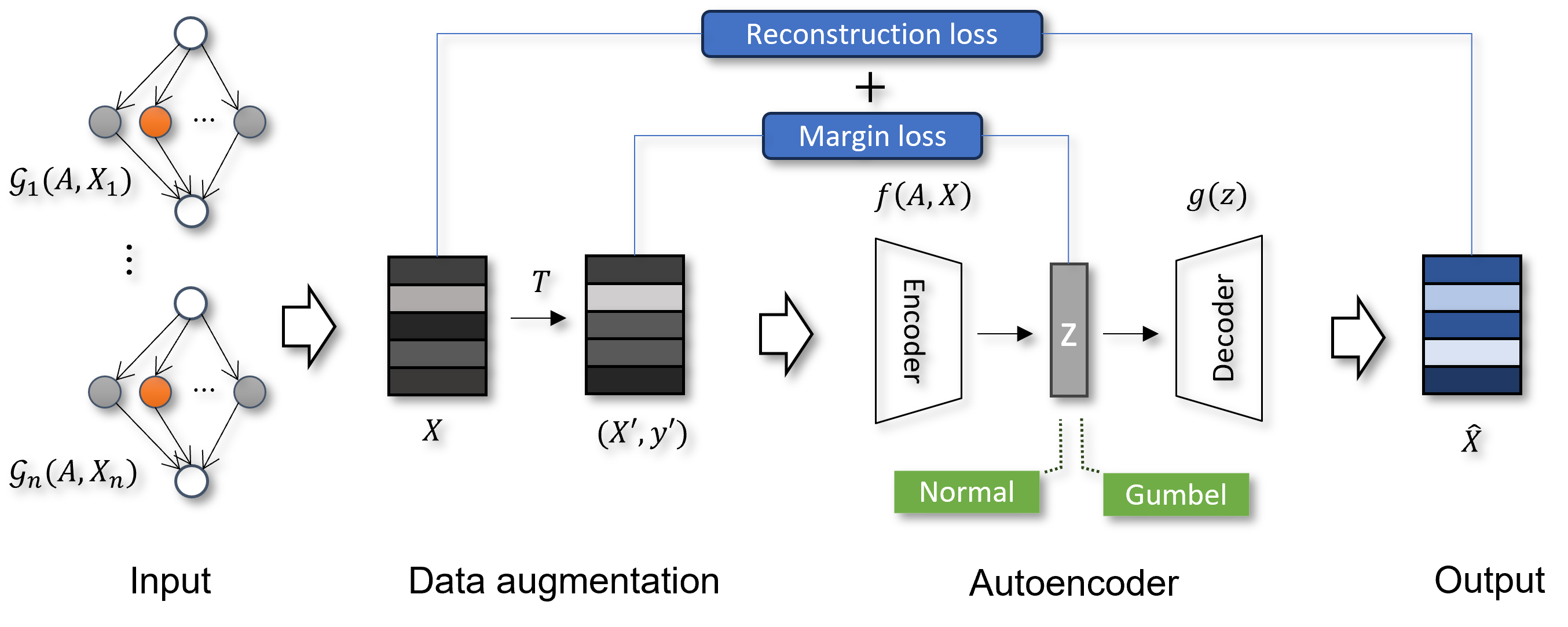}
  \caption{Self-supervised learning: inputs are a set of DAGs representing the scientific workflows $\Gcal(\Avec, \Xvec)$. The outputs are the matrices representing the reconstructed feature matrices $\hat{\Xvec}$.
  }
  \label{fig:ssl}
\end{figure*}
Self-supervised learning is a type of unsupervised learning that creates target objectives without human annotation. It has led to a dramatic increase in the ability to capture salient feature representations from unlabeled visual data \cite{reed2021selfaugment}. A common practice in self-supervised learning is to perform data augmentation via hand-crafted transformations intended to leave the semantics of the data invariant~\cite{von2021self}. Recent works have used extensive supervised evaluations to determine which augmentations are most effective~\cite{reed2021selfaugment}.

The self-supervised learning setup involves two key steps. The first step is to perform data augmentation and to generate a modified version of the input graph through transformation. The second step is to use both the transformed input graph and the original graph and learn the latent space. We will first describe the data augmentation procedure utilized to generate these transformations.

\subsubsection{Data Augmentation}
% \note{revised.}
% {\color{blue}
Data augmentation is a crucial component of self-supervised learning on graphs, which enables to increase the diversity of the training data and to encourage the model to learn more generalizable representations.
By applying various augmentation techniques, such as node and edge perturbations, graph augmentation can simulate different scenarios and generate new graph samples that are similar to the original data but not identical.
This helps to prevent overfitting and improves the model's ability to generalize to unseen data.
Moreover, data augmentation can also help to reduce the risk of over-smoothing, which occurs when a model becomes too specialized to the training data and fails to capture the underlying structure of the data.
By generating new data samples that capture different aspects of the graph structure, data augmentation can encourage the model to learn more robust and diverse representations.
% }

To be precise, let $\Xvec, \Yvec$ be the input data and target output of the primary task, respectively.
And let $\Xvec'$ be the augmented attributed matrix~(the transformed data), with $\Xvec' \in \RR^{n \times d}$ that is generated by performing feature augmentation and modifications on $\Xvec.$  The modified data is obtained by applying a transformation $T$ to the original input data $\Xvec' = T(\Xvec)$.
The transformation $T$ is designed to preserve the semantic content of the input data while introducing variations that are useful for data augmentation.
A common method of doing this is to evaluate an average deviation of the feature values at a particular node with its neighbors~\cite{song2022graph}.

To calculate this, we first select a group of nodes from the graph according to an arbitrary probability distribution. Given $\pvec$ to be a random probability of selecting nodes, we select a group of nodes through a mask given as
\begin{align}
  \label{eq:mask}
  \mvec = (\pvec \ge \frac{1}{2}r) \land (\pvec \le \frac{3}{4}r),
\end{align}
where $r$ is a selection rate between $\sbr{0, 1}$. Applying the mask on the graph provides a selection of nodes. Next, for each node in this selection, we evaluate the pairwise distance between a node and its neighbors. The pairwise distances are denoted as $d_{u, v}, v \in \Ncal(u)$, where $v$ is in the neighbor set of $u$. Then, based on the minimum distance $d_{MIN} = min_{v \in \Ncal(u)} d_{u, v}$, we select the features of the closest neighbor and augment the original vector $\Xvec$ to obtain the augmented feature vector $\Xvec'$.

Next, we select a factor and scale the features. Given a pre-selected factor, we obtain two groups of nodes by randomly generating two masks according to \eqref{eq:mask}. For the first group of nodes, we scale up the set of features and scale down the set of features for the second group. Utilizing these two augmentation procedures, we obtain a new set of feature matrix $\Xvec'$ and define a set of augmented labels as
\begin{align}
  \yvec' = \bI_{\pvec < r},
\end{align}
where $\bI$ is the indicator function. Given the original and augmented data vector, we apply an encoder and obtain the latent space.

\subsubsection{Encoder}
One example of an encoder in the SSL context is a masked autoencoder~\cite{he2022masked}, in which the random patches of the input image are masked and the missing pixels are reconstructed by the autoencoder. We utilize the masked autoencoder structure but replace the convolutional neural network with GraphSAGE~\cite{hamilton2017inductive}. GraphSAGE layers utilize a convolutional operation to aggregate information from neighboring nodes in the graph. The aggregated information is then passed through a nonlinear activation function to produce a lower-dimensional representation of the input data.
Using the encoder, we will transform both the augmented data and the original data into the latent space where we can evaluate the summary statistics. We denote these transformations as $f(\Xvec, \Avec)$, and $f(T(\Xvec), \Avec).$

\subsection{Step 2: Latent Space}
We denote the latent space by $\zvec$ and obtain a hidden layer representation by sampling from $f(\Xvec, \Avec)$ and $f(T(\Xvec), \Avec)$ to obtain $\zvec$ and $\zvec'.$  Here we advocate two approaches for sampling $z$ and $z'$, first through the normal distribution and then through the Gumbel distribution.

\subsubsection{Normal Distribution}
We sample from the normal distribution at the latent space by the reparameterization trick~\cite{kingma2019introduction}. It involves rewriting the sampling operation so that it is differentiable with respect to the parameters of the distribution. Specifically, instead of sampling directly from the distribution, we sample from a standard normal distribution and then transform the samples using the mean and variance of the distribution. This allows us to backpropagate through the sampling operation and optimize the parameters of the encoder network using stochastic gradient descent. Formally, we express the sampled normal distribution as
\begin{align}
  z(\Xvec, \Avec) = \muvec + \sigmavec \odot \vec{\epsilon},
\end{align}
where $\muvec$ and $\sigmavec$ are the mean and standard deviation of the normal distribution parameterized in the hidden space, respectively,  and $\vec{\epsilon} \sim \Ncal (\zero, \one)$ is a sampled normal distribution with zero mean and unit variance.

\subsubsection{Gumbel Distribution}
The idea is that the data points corresponding to the abnormal behavior are captured on the tail of the distribution and the normal behavior is captured at the center of the distribution. In probability terms, this can be interpreted as normal behavior should be captured with probability values close to one and abnormal behavior close to zero. Therefore we would like to model a continuous function between zero and one with a peak at one and a bottom at zero. This behavior is appropriately captured by the Gumbel distribution.

Gumbel distributions belong to a class of generalized extreme value distribution utilized to capture abnormal behavior~\cite{cooray2010generalized}.  Recently, the Gumbel softmax trick was introduced to enforce a Gumbel distribution~\cite{huijben2022review} at the output of the encoder. In the Gumbel softmax trick, a continuous approximation of a bivariate discrete distribution is obtained, and it is ensured that  the values zero and one are prioritized by the latent space, namely, a sharp peak at one and an extended tail near zero with the key property that one should be able to sample from this space.

To sample from the Gumbel distribution at the latent space, we utilize the Gumbel softmax trick. In the Gumbel softmax trick, the value of the hidden layer representation is provided by
\begin{align}
  z(\Xvec, \Avec) & = \frac{ \rho + f(\Xvec, \Avec) }{t}                    \\
  z(\Xvec, \Avec) & = z(\Xvec, \Avec) - max(z(\Xvec, \Avec))                \\
  z(\Xvec, \Avec) & = \frac{e^{z(\Xvec, \Avec)}}{\sum e^{z(\Xvec, \Avec)}},
\end{align}
where $\rho \sim log(-log(p + \epsilon)),$ where $p \sim \mathcal{U}(0,1)$ and $\epsilon \in \RR.$ Moreover, $t$ controls the shape of the Gumbel distribution. By sampling $\epsilon$ and $\rho$ from a uniform distribution and combining it with the latent space representation of both $\Xvec$ and $\Xvec',$ we can generate samples from the latent space and appropriately reconstruct the feature matrix through the decoder. We utilize a multilayer perceptron for the decoder.

\subsection{Step 3: Multilayer Perceptron as a Decoder}
The goal of the decoder is to reconstruct the original feature matrix from the lower-dimensional representation produced by the encoder \cite{chen2022context}. The decoder is typically a mirror image of the encoder, with each layer performing an inverse operation to that performed by its corresponding layer in the encoder. However, since, we seek to reconstruct just the feature matrix from the graph data, we will simply utilize a multilayer perceptron to perform this step. In this scenario, we will write
\begin{align}
  \hat{\Xvec} =f(z(\Xvec, \Avec)).
\end{align}
Note that the augmented data vector is required only  to construct a rich latent space and therefore the MLP is applied only to the latent layer representation of the original graph.
Once a reconstruction is obtained, we have the output of our model.
\subsection{Loss Function and the Overall Setup}
The model is trained to predict an anomaly score on the original data. To this end, we assume that all the data we have available for analysis  belongs to the normal class. Therefore, the goal of training is to get as close as possible to the original feature matrix in the reconstruction of the modified data; any deviation from the original feature matrix indicates an abnormality. For this training, we will define the following loss function such that
\begin{align}
  \mathcal{L}= \eta L_{\text{reconstruction error} } + (1-\eta) L_{\text{margin} },
\end{align}
where we define the reconstructed loss as
\begin{align}
  L_{\text{reconstruction error}} (\hat{\Xvec}, \Xvec) = \| \hat{\Xvec} - \Xvec \|.
\end{align}
Similarly, we apply the margin loss from the augmented label as
\begin{align}
  L_{\text{margin}} = \max(0, - \yvec' (\zvec - \zvec') + \lambda),
\end{align}
where $\lambda$ is a predefined margin value. Specific to a single job $i$, we can obtain the anomaly score as
\begin{align}
  \label{eq:anomaly_score}
  \svec_i = \eta \| \hat{\Xvec}_i - \Xvec_i \|  + (1-\eta) \max(0, - \yvec_i' (\zvec_i - \zvec_i') + \lambda).
\end{align}
This allows us to quantify the significance of a threat job in the workflow. Once the model is trained to reconstruct the input data, the anomaly score can be used as a measure of anomaly. An input data point is considered anomalous if its anomaly score is significantly higher than the anomaly scores of the normal data points. The anomaly score can be thresholded to identify anomalous data points:
\begin{align}
  \tilde{y}(\Xvec_i, \Avec) =
  \begin{cases}
    1 \quad & \text{if } \svec_i > \tau \\
    0 \quad & \text{otherwise}.
  \end{cases}
\end{align}
Here $\tau$ is a threshold determined by the user.

\section{Experiments}

\subsection{Dataset}
The experimental data for our models is available through Flow-Bench~\cite{papadimitriou2023flowbench}, a benchmark dataset for anomaly detection in science workflows. Flow-Bench contains workflow execution traces of two computational science workflows and one data science workflow, executed on a distributed infrastructure, using the Pegasus Workflow Management System ~\cite{deelman-fgcs-2015}. This dataset captures $1,211$ DAG executions of the Pegasus 1000Genome workflow, the Pegasus Montage workflow, and the Pegasus Predict Future Sales workflow, under normal conditions, as well as conditions with synthetically injected anomalies. The synthetic anomalies are injected on a per-node level, where each DAG execution has only a single type of anomaly. Table~\ref{tab:dataset} provides details about the number of DAG executions per anomaly class. We introduce two major anomaly classes: (1) CPU anomaly ($CPU\_K$), where $K$ cores are marked as unavailable on some workers; and (2) Disk anomaly ($HDD\_K$), where the average write speed to the disk is capped at $K$ MB/s and the read speed is capped at $(2\times K)$ MB/s. Table~\ref{tab:dataset} summarizes the three workflows.

\begin{table}[h]
  \centering
  \caption{Dataset Description.}
  {
    \label{tab:dataset}
    \resizebox{\linewidth}{!}{
      \begin{tabular}{l||cc|c|ccc|cc}
        \toprule
                                  & \multicolumn{2}{c|}{DAG Information}
                                  & \multicolumn{6}{c}{\#DAG Executions}                                                                      \\
        \cmidrule{2-9}
        \multirow{2}{*}{Workflow} & \multirow{2}{*}{Nodes}               & \multirow{2}{*}{Edges}
                                  & \multirow{2}{*}{Normal}              & \multicolumn{3}{c|}{CPU} & \multicolumn{2}{c}{HDD}                 \\
                                  &                                      &
                                  &                                      & 2                        & 3                       & 4  & 5   & 10 \\
        \midrule
        1000 Genome               & 137                                  & 289
                                  & 51                                   & 100                      & 25                      & -  & 100 & 75 \\
        Montage                   & 539                                  & 2838
                                  & 51                                   & 46                       & 80                      & -  & 67  & 76 \\
        Predict Future Sales      & 165                                  & 581
                                  & 100                                  & 88                       & 88                      & 88 & 88  & 88 \\
        \bottomrule
      \end{tabular}
    }}
\end{table}

\subsubsection{1000 Genome Workflow}
\label{subsubsec:1000genome-workflow}
\begin{figure}[h]
  \centering
  \includegraphics[width=\linewidth]{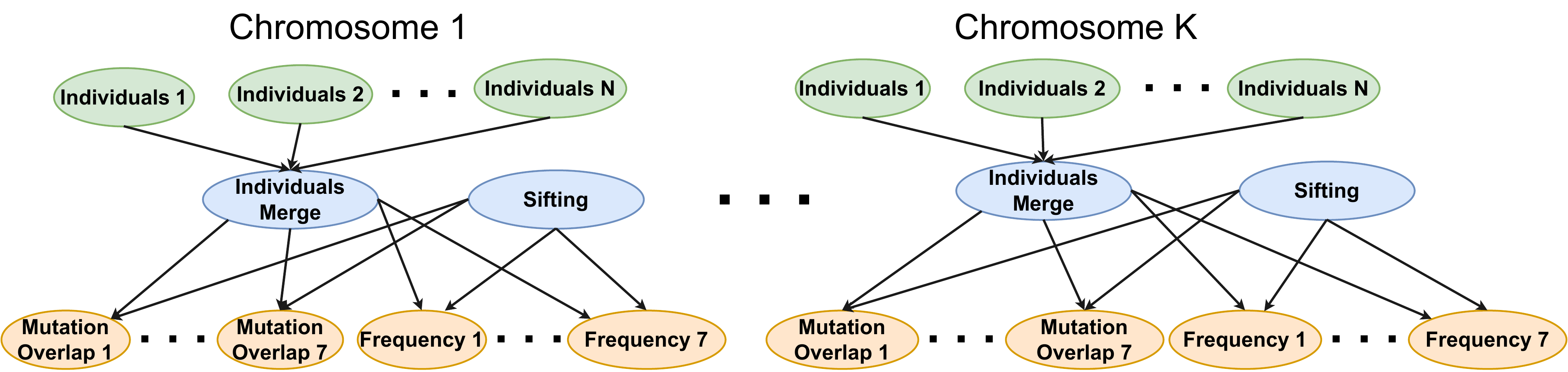}
  \caption{\footnotesize Overview of the 1000 Genomes sequencing analysis workflow. The workflow creates a branch for each chromosome, and each individual task is processing a subset of the Phase 3 data (equally distributed).}
  \label{fig:1000genome}
  \vspace{-10pt}
\end{figure}

The 1000 Genomes Project~\cite{1000genome-project} provides a reference for human variation, having reconstructed the genomes of 2,504 individuals across 26 different populations. The test case we have here identifies mutational overlaps using data from the 1000 Genomes Project in order to provide a null distribution for rigorous statistical evaluation of potential disease-related mutations. The workflow (Figure~\ref{fig:1000genome}) is composed of five  tasks that calculate the mutation overlap and the frequency of overlapping mutations for six populations (African, Mixed American, East Asian, European, South Asian, and British).

\subsubsection{Montage Workflow}
\label{subsubsec:montage-workflow}
\begin{figure}[h]
  \centering
  \includegraphics[width=\linewidth]{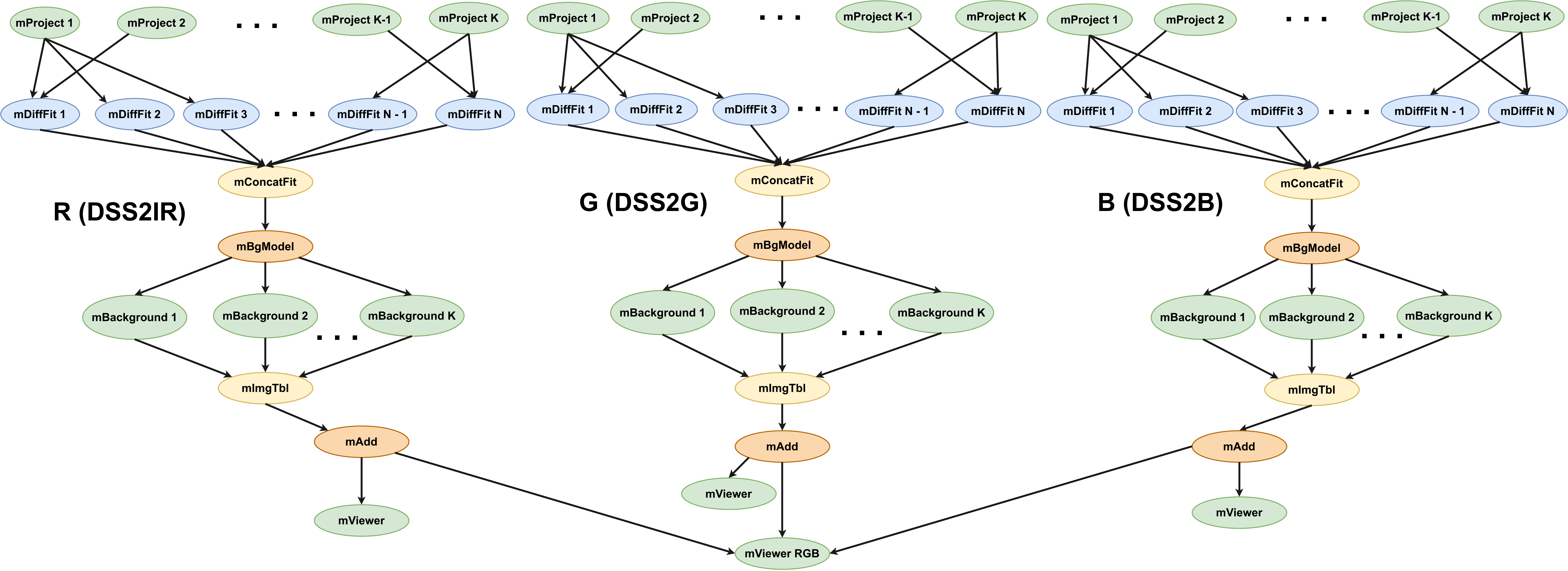}
  \caption{\footnotesize Overview of the Montage workflow. In this case, the workflow uses images captured by the Digitized Sky Survey (DSS)~\cite{dss-archive} and creates a branch for each band that is requested to be processed during the workflow generation. The size of the first level of each branch depends on the size of the section of the sky to be analyzed, while the second level depends on the number of overlapping images stored in the archive.}
  \label{fig:montage}
\end{figure}

Montage is an astronomical image toolkit~\cite{2010ascl.soft10036J} with components for reprojection, background matching, coaddition, and visualization of FITS files. Montage workflows typically follow a predictable structure based on the inputs, with each stage of the workflow often taking place in discrete levels separated by some synchronization/reduction tasks (mConcatFit and mBgModel). The workflow (Figure~\ref{fig:montage}) uses  Montage to transform astronomical images into custom mosaics;  the total data footprint can range to many gigabytes.

\subsubsection{Predict Future Sales Workflow}
\label{subsubsec:predict-future-sales-workflow}
\begin{figure}[h]
  \centering
  \includegraphics[width=1\linewidth]{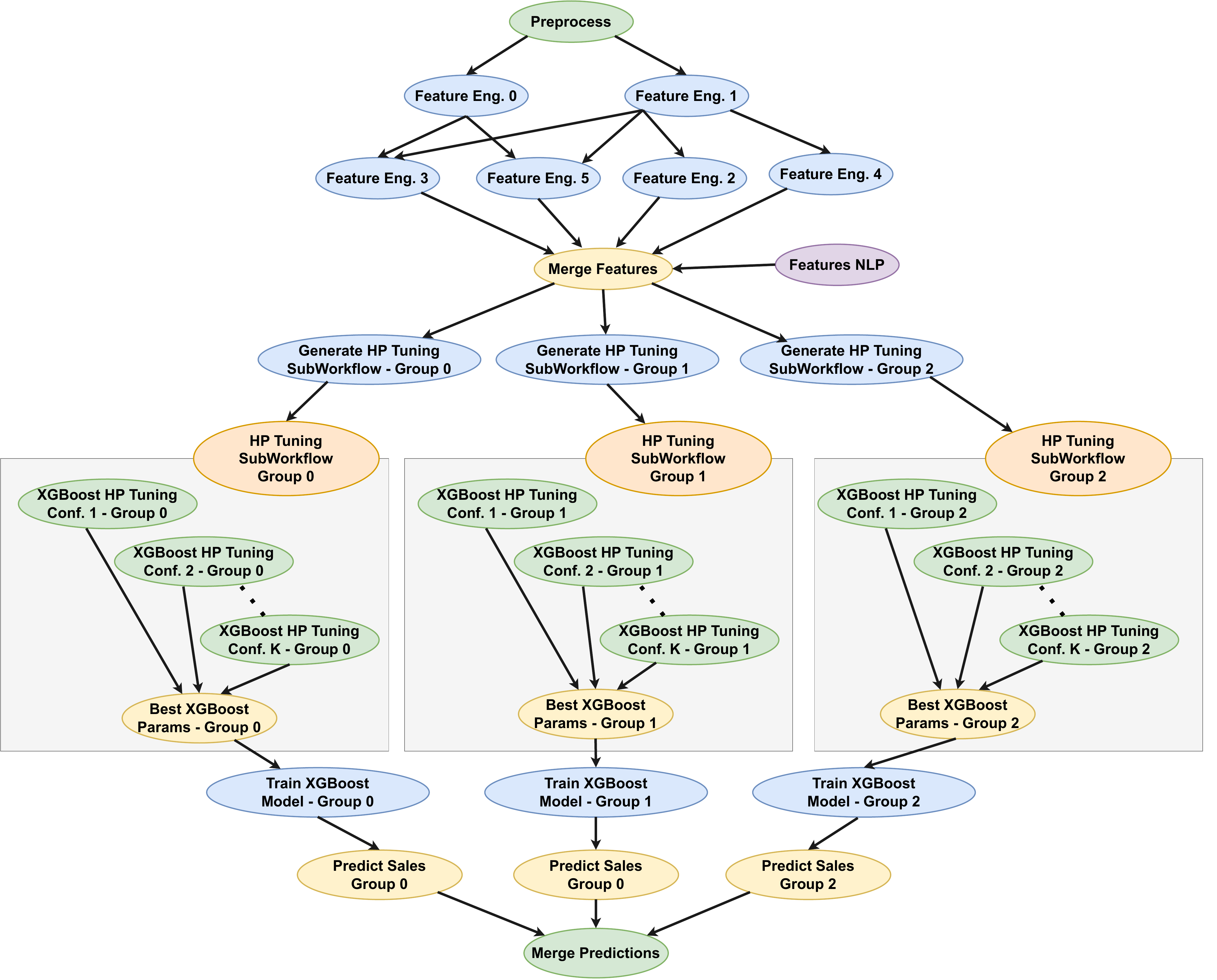}
  \caption{\footnotesize Overview of the Predict Future Sales workflow. The workflow splits the data into 3 item categories and trains 3 XGBoost models that are later combined, using an ensemble technique. It contains 3 hyperparameter tuning sub-workflows that test different sets of features and picks the best-performing one. The number of HPO tasks is configurable and depends on the number of combinations that will be tested.}
  \label{fig:predict-future-sales}
  \vspace{-10pt}
\end{figure}

The Predict Future Sales data science workflow provides a solution to Kaggle’s predict future sales competition~\cite{kaggle-predict-future-sales}. The workflow receives daily historical sales data from January 2013 to October 2015 and attempts to predict the sales for November 2015. The workflow (Figure~\ref{fig:predict-future-sales}) includes multiple preprocessing and feature engineering steps to augment the dataset with new features and separates the dataset into three major groups based on their type and sales performance. To improve the prediction score, the workflow goes through a hyperparameter tuning phase and trains three discrete XGBoost models for each item group. In the end, it applies a simple ensemble technique that uses the appropriate model for each item prediction and combines the results into a single output file.

For a more detailed description of the workflows, please refer to Flow-Bench~\cite{papadimitriou2023flowbench}.
%, which is composed of eleven different tasks: (1) Preprocess -- performs data cleanup and removes outliers; (2) Feature Eng. X -- each of these tasks extends the dataset by creating new features, using the currently available ones; (3) Features NLP -- creates word embeddings and adds new features that can be used to cluster similar items together, based on their names; (4) Merge Features -- merges all features, splits the items into 3 distinct groups and separates the dataset into training and test sets; (5) Generate HP Tuning SubWorkflow -- generates a hyperparameter tuning subworkflow, where each task is testing a different set of features; (6) HP Tuning Workflow -- submits and monitors the execution of the subworkflow; (7) XGBoost HP Tuning Conf. X -- each of these tasks considers a different set of features and uses Hyperopt~\cite{bergstra2013making} to tune the XGBoost~\cite{chen2015xgboost} model parameters, the number of tasks is configurable based on how many feature combinations we want to test; (8) Best XGBoost Params -- compares the results of each hyperparameter tuning round and picks the best parameters; (9) Train XGBoost Model -- trains the XGBoost model with the best parameters found; (10) Predict Sales -- performs the sales prediction using the trained XGBoost model; (11) Merge Predictions -- combines the predictions of all groups together in a single file.

\subsection{Experimental Setup}
Flow-Bench~\cite{papadimitriou2023flowbench} provides detailed instructions to process the workflow data as a dataset in Pytorch-Geometric~\cite{Fey/Lenssen/2019} with shifted timestamps and column-wise normalization over jobs. We denoted label vector $\yvec$ to indicate job anomalies, where $y_i=1$ if job $i$ was an anomaly and $y_i=0$ otherwise. Those labels were not used in the training stage, but only for evaluation.
We further split the dataset into training, validation, and test set with a ratio $0.6$, $0.2$, and $0.2$, respectively,  and used the data loader to load the dataset in mini-batches for efficient training.
We compared our SSL model with several unsupervised learning methods from PyOD~\cite{zhao2019pyod} and PyGOD~\cite{liu2022bond}, which are the benchmarks of anomaly detection for tabular-based and graph-based data, respectively. For each method, we used the same dataset of workflows and applied the same preprocessing steps, such as splitting, sampling, and normalization. Without specific mention, we use the default parameters of each method. For a detailed description of each individual benchmark method, please refer to the Flow-Bench ~\cite{papadimitriou2023flowbench}.

\subsubsection{Metrics}
As we focus on unsupervised learning, we use three metrics to measure the performance of our model and the baselines: ROC-AUC score, average precision score, and top-k precision. ROC-AUC score is the area under the receiver operating characteristic curve, which reflects how well the model can distinguish between different classes. The average precision score is the average of the precision scores for each class, which indicates whether one's model can correctly identify all the positive examples without accidentally marking too many negative examples as positive. Top-k precision is the proportion of times when the correct label is among the top k labels predicted by the model. The top-k evaluation method is widely used in anomaly detection to evaluate the performance of an algorithm~\cite{petersen2022differentiable}. It involves ranking the anomalies in a dataset based on their scores and then selecting the top-k anomalies for evaluation.

\subsubsection{Training}
\begin{figure}
  \centering
  \includegraphics[width=0.7\linewidth]{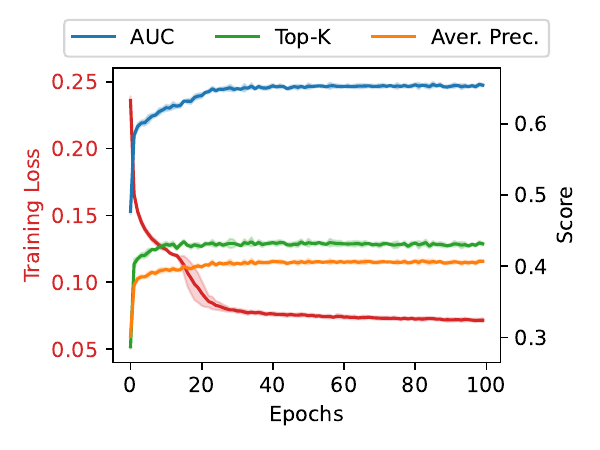}
  \caption{Training of SSL (1000 Genomes Project)}
  \label{fig:training}
  \vspace{-10pt}
\end{figure}

We use the Adam optimizer with a learning rate of $1e^{-3}$ and a weight decay of $1e^{-4}$, and we train the model for 100 epochs with a batch size of 32. The model is efficient to fit into a single GPU setting with only about 1.5 GB of memory consumption.
Figure~\ref{fig:training} reports the training process of our SSL model on the 1000 Genomes workflow with 5 independent runs, in which the red curve is the training loss while the others are the metrics we measured for our model.
Notice that the training loss drops rapidly in the first few epochs of training, which indicates the rapid convergence of our model. Also, with just about 30 epochs, the model can achieve  stable performance, which is a significant advantage of our model compared with other baselines.

\subsubsection{Sensitivity Analysis of Hyperparameters}
Similar to other deep learning models, our SSL model involves a set of hyperparameters that are sensitive for training and evaluation.
Those hyperparameters include the learning rate, the batch size, the number of epochs, and the regularization coefficient.
Specific to our SSL model, we have a set of hyperparameters that play key roles in the anomaly score, such as $r$ and $\tau$, the selection rate for augmented data and the threshold for anomalous pseudo labels, respectively.
To investigate the sensitivity of those hyperparameters, we tuned the hyperparameters using DeepHyper~\cite{balaprakash2018deephyper}, a powerful package for automating machine learning hyperparameter tuning based on Bayesian optimization.
We predefined the search domain for hyperparameters including hidden dimension, number of layers, learning rate, the decay rate for Adam optimizer, dropout rate, number of neighbors for data loader, selection rate, and the threshold of the pseudo label. We targeted a single objective of loss on the validation set, and we report the best-performed model in Table~\ref{tab:model_comp_unsupervised}.

\subsection{Experiment results}

Table~\ref{tab:model_comp_unsupervised} presents the impressive results of our approach to the three workflows. In general, our approach outperformed both tabular-based and graph-based benchmark methods. Our SSL models while utilizing both normal and Gumbel distributions, achieved better performance with high efficiency in handling large workflows, without encountering out-of-memory (OOM) or time-limit-exceed (TLE) errors. In contrast, many deep learning methods encountered issues with OOM and TLE due to the inefficient handling of large workflow sizes, particularly in graph completion and reconstruction methods~\cite{yuan2021higher, xu2022contrastive}.

Note that the deep learning methods exhibit a slightly better performance in reported metrics compared with the tabular-based methods in the middle of the table albeit with a large variance in the metrics. The only exception to the superior performance of our method is the KNN for the Predict Future Sales data science workflow where KNN achieved a slightly better performance under the measure of top-k precision scores: KNN is approximately 0.06 better than our approach. For the other two works, however, KNN exhibited poorer performance relative to our approach.
Overall, our results demonstrate that our SSL models are highly effective in handling large workflows and achieving superior performance, making them a promising approach for various practical applications.

\iffalse
  \begin{table}[t]
    \centering
    \caption{Training time per epoch}
    \label{tab:time_efficiency}
    \begin{tabular}{c|c|c|c}
      \toprule
      Workflow                & \# of nodes & w/o partition & w/ partition \\
      \midrule
      1000 Genome             & 137         & 0.78          & 0.23         \\
      Montage                 & 539         & 1.54          & 0.75         \\
      Future Sales Prediction & 165         & 0.86          & 0.41         \\
      \bottomrule
    \end{tabular}
    \vspace{-10pt}
  \end{table}
\fi

\begin{table*}[h]
  \centering
  \caption{Model comparison. Reported the ``mean $\pm$ standard deviation (maximum)'' of the metrics across five runs for each method.}
  \label{tab:model_comp_unsupervised}
  \resizebox{\textwidth}{!}{
    \begin{tabular}{r|ccc|ccc|ccc}
      \toprule
                  & \multicolumn{3}{c|}{1000 Genome}
                  & \multicolumn{3}{c|}{Montage}
                  & \multicolumn{3}{c}{Predict Future Sales}                                                               \\
      \cmidrule{2-10}
                  & ROC-AUC                                  & Ave. Prec.                   & Prec. @ k
                  & ROC-AUC                                  & Ave. Prec.                   & Prec. @ k
                  & ROC-AUC                                  & Ave. Prec.                   & Prec. @ k                    \\
      \midrule
      MLPAE       & .545 $\pm$ .012 (.593)                   & .356 $\pm$ .026 (.412)       & .430 $\pm$ .017 (.451)
                  & .518 $\pm$ .007 (.532)                   & .208 $\pm$ .021 (.271)       & .198 $\pm$ .014 (.215)
                  & .508 $\pm$ .009 (.513)                   & .120 $\pm$ .010 (.131)       & .153 $\pm$ .004 (.161)       \\
      SCAN        & .491 $\pm$ .005 (.497)                   & .323 $\pm$ .002 (.341)       & .274 $\pm$ .004 (.281)
                  & .500 $\pm$ .000 (.500)                   & .204 $\pm$ .001 (.204)       & .230 $\pm$ .001 (.230)
                  & .500 $\pm$ .000 (.500)                   & .104 $\pm$ .001 (.104)       & .000 $\pm$ .000 (.000)       \\
      Radar       & \multicolumn{3}{c|}{OOM}
                  & \multicolumn{3}{c|}{OOM}
                  & .632 $\pm$ .011 (.645)                   & .158 $\pm$ .003 (.162)       & .191 $\pm$ .010 (.194)       \\
      Anomalous   & \multicolumn{3}{c|}{OOM}
                  & \multicolumn{3}{c|}{OOM}
                  & .607 $\pm$ .021 (.612)                   & .176 $\pm$ .009 (.177)       & .210 $\pm$ .007 (.225)       \\
      GCNAE       & .610 $\pm$ .017 (.620)                   & .380 $\pm$ .005 (.386)       & .426 $\pm$ .004 (.430)
                  & .519 $\pm$ .009 (.523)                   & .211 $\pm$ .003 (.219)       & .146 $\pm$ .000 (.146)
                  & .398 $\pm$ .020 (.405)                   & .098 $\pm$ .000 (.098)       & .125 $\pm$ .008 (.135)       \\
      Dominant    & \multicolumn{3}{c|}{OOM}
                  & \multicolumn{3}{c|}{OOM}
                  & .427 $\pm$ .018 (.432)                   & .089 $\pm$ .000 (.089)       & .112 $\pm$ .006 (.113)       \\
      DONE        & \multicolumn{3}{c|}{OOM}
                  & \multicolumn{3}{c|}{OOM}
                  & .559 $\pm$ .016 (.567)                   & .130 $\pm$ .001 (.131)       & .143 $\pm$ .008 (.146)       \\
      ADONE       & \multicolumn{3}{c|}{OOM}
                  & \multicolumn{3}{c|}{OOM}
                  & .579 $\pm$ .010 (.586)                   & .147 $\pm$ .010 (.151)       & .166 $\pm$ .007 (.169)       \\
      AnomalyDAE  & \multicolumn{3}{c|}{OOM}
                  & \multicolumn{3}{c|}{OOM}
                  & .584 $\pm$ .022 (.595)                   & .146 $\pm$ .003 (.155)       & .145 $\pm$ .003 (.148)       \\
      GAAN        & \multicolumn{3}{c|}{OOM}
                  & \multicolumn{3}{c|}{OOM}
                  & .618 $\pm$ .014 (.629)                   & .160 $\pm$ .008 (.169)       & .175 $\pm$ .001 (.178)       \\
      GUIDE       & \multicolumn{3}{c|}{TLE}
                  & \multicolumn{3}{c|}{TLE}
                  & \multicolumn{3}{c}{TLE}                                                                                \\
      CONAD       & \multicolumn{3}{c|}{TLE}
                  & \multicolumn{3}{c|}{TLE}
                  & \multicolumn{3}{c}{TLE}                                                                                \\
      \midrule
      ABOD        & .500 $\pm$ .009 (.510)                   & .326 $\pm$ .009 (.333)       & .249 $\pm$ .003 (.258)
                  & .500 $\pm$ .003 (.510)                   & .204 $\pm$ .005 (.209)       & .230 $\pm$ .000 (.230)
                  & .500 $\pm$ .008 (.505)                   & .104 $\pm$ .005 (.110)       & .000 $\pm$ .001 (.001)       \\
      CBLOF       & .489 $\pm$ .007 (.500)                   & .322 $\pm$ .003 (.324)       & .346 $\pm$ .010 (.355)
                  & .517 $\pm$ .004 (.523)                   & .211 $\pm$ .000 (.211)       & .233 $\pm$ .004 (.238)
                  & .546 $\pm$ .004 (.550)                   & .119 $\pm$ .003 (.128)       & .182 $\pm$ .000 (.182)       \\
      FB          & .509 $\pm$ .001 (.518)                   & .330 $\pm$ .008 (.331)       & .297 $\pm$ .000 (.297)
                  & .487 $\pm$ .010 (.488)                   & .201 $\pm$ .002 (.201)       & .205 $\pm$ .005 (.215)
                  & .451 $\pm$ .003 (.458)                   & .103 $\pm$ .009 (.105)       & .006 $\pm$ .004 (.014)       \\
      HBOS        & .513 $\pm$ .008 (.515)                   & .333 $\pm$ .003 (.333)       & .365 $\pm$ .001 (.370)
                  & .570 $\pm$ .009 (.571)                   & .253 $\pm$ .006 (.257)       & .302 $\pm$ .000 (.302)
                  & .539 $\pm$ .009 (.546)                   & .116 $\pm$ .005 (.121)       & .169 $\pm$ .003 (.169)       \\
      IF          & .504 $\pm$ .000 (.513)                   & .328 $\pm$ .009 (.336)       & .357 $\pm$ .008 (.357)
                  & .497 $\pm$ .005 (.507)                   & .204 $\pm$ .003 (.206)       & .209 $\pm$ .007 (.217)
                  & .560 $\pm$ .002 (.568)                   & .127 $\pm$ .001 (.134)       & .207 $\pm$ .009 (.214)       \\
      KNN         & .558 $\pm$ .003 (.566)                   & .374 $\pm$ .007 (.380)       & .384 $\pm$ .001 (.393)
                  & .566 $\pm$ .008 (.567)                   & .250 $\pm$ .006 (.256)       & .297 $\pm$ .005 (.300)
                  & .663 $\pm$ .005 (.664)                   & .232 $\pm$ .002 (.232)       & {\bf .382 $\pm$ .007 (.395)} \\
      AKNN        & .547 $\pm$ .001 (.547)                   & .366 $\pm$ .005 (.366)       & .356 $\pm$ .010 (.359)
                  & .553 $\pm$ .000 (.556)                   & .243 $\pm$ .009 (.244)       & .286 $\pm$ .006 (.295)
                  & .642 $\pm$ .004 (.645)                   & .224 $\pm$ .007 (.231)       & .327 $\pm$ .009 (.330)       \\
      LOF         & .509 $\pm$ .005 (.515)                   & .331 $\pm$ .007 (.335)       & .297 $\pm$ .005 (.307)
                  & .492 $\pm$ .007 (.495)                   & .202 $\pm$ .003 (.202)       & .221 $\pm$ .005 (.231)
                  & .480 $\pm$ .008 (.481)                   & .102 $\pm$ .002 (.106)       & .058 $\pm$ .001 (.064)       \\
      MCD         & .503 $\pm$ .005 (.511)                   & .328 $\pm$ .005 (.334)       & .355 $\pm$ .004 (.357)
                  & .514 $\pm$ .006 (.521)                   & .210 $\pm$ .000 (.210)       & .230 $\pm$ .005 (.239)
                  & .510 $\pm$ .005 (.520)                   & .106 $\pm$ .002 (.107)       & .117 $\pm$ .002 (.118)       \\
      OCSVM       & .510 $\pm$ .003 (.513)                   & .331 $\pm$ .004 (.337)       & .355 $\pm$ .004 (.362)
                  & .464 $\pm$ .003 (.467)                   & .199 $\pm$ .000 (.199)       & .146 $\pm$ .010 (.151)
                  & .540 $\pm$ .007 (.546)                   & .117 $\pm$ .005 (.122)       & .172 $\pm$ .001 (.173)       \\
      PCA         & .500 $\pm$ .007 (.508)                   & .326 $\pm$ .008 (.330)       & .249 $\pm$ .008 (.249)
                  & .500 $\pm$ .004 (.505)                   & .204 $\pm$ .000 (.204)       & .230 $\pm$ .010 (.238)
                  & .529 $\pm$ .000 (.529)                   & .112 $\pm$ .006 (.116)       & .152 $\pm$ .009 (.156)       \\
      LSCP        & .494 $\pm$ .003 (.495)                   & .324 $\pm$ .010 (.327)       & .289 $\pm$ .000 (.289)
                  & .490 $\pm$ .003 (.500)                   & .202 $\pm$ .008 (.205)       & .206 $\pm$ .003 (.209)
                  & .530 $\pm$ .004 (.536)                   & .113 $\pm$ .008 (.120)       & .159 $\pm$ .009 (.161)       \\
      INNE        & .503 $\pm$ .002 (.511)                   & .328 $\pm$ .006 (.335)       & .350 $\pm$ .008 (.351)
                  & .461 $\pm$ .006 (.466)                   & .200 $\pm$ .005 (.205)       & .145 $\pm$ .000 (.145)
                  & .466 $\pm$ .009 (.474)                   & .102 $\pm$ .008 (.105)       & .030 $\pm$ .008 (.033)       \\
      GMM         & .486 $\pm$ .003 (.489)                   & .321 $\pm$ .001 (.322)       & .341 $\pm$ .004 (.346)
                  & .502 $\pm$ .009 (.506)                   & .205 $\pm$ .005 (.209)       & .203 $\pm$ .008 (.212)
                  & .532 $\pm$ .001 (.542)                   & .113 $\pm$ .004 (.116)       & .157 $\pm$ .004 (.166)       \\
      KDE         & .514 $\pm$ .007 (.516)                   & .334 $\pm$ .009 (.337)       & .357 $\pm$ .001 (.365)
                  & \multicolumn{3}{c|}{TLE}
                  & .491 $\pm$ .006 (.500)                   & .103 $\pm$ .004 (.105)       & .084 $\pm$ .005 (.088)       \\
      LMDD        & \multicolumn{3}{c|}{TLE}
                  & \multicolumn{3}{c|}{TLE}
                  & .557 $\pm$ .011 (.566)                   & .125 $\pm$ .005 (.131)       & .202 $\pm$ .003 (.203)       \\
      \midrule
      SSL(Normal) & .675 $\pm$ .021 (.690)                   & .405 $\pm$ .007 (.412)       & .432 $\pm$ .007 (.440)
                  & .591 $\pm$ .015 (.624)                   & .251 $\pm$ .001 (.255)       & {\bf .297 $\pm$ .007 (.302)}
                  & {\bf .687 $\pm$ .013 (.698)}             & .245 $\pm$ .008 (.245)       & .345 $\pm$ .006 (.364)       \\
      SSL(Gumbel) & {\bf .689 $\pm$ .016 (.712)}             & {\bf .475 $\pm$ .004 (.477)} & {\bf .488 $\pm$ .009 (.505)}
                  & {\bf .610 $\pm$ .009 (.634)}             & {\bf .265 $\pm$ .007 (.274)} & .259 $\pm$ .001 (.262)
                  & .681 $\pm$ .009 (.689)                   & {\bf .271 $\pm$ .004 (.280)} & .382 $\pm$ .001 (.389)       \\
      \bottomrule
    \end{tabular}
  }
\end{table*}

\subsubsection{Ranking by Anomaly Scores}
In addition to the impressive performance, our approach enables retrieval of an anomaly score summarizing the degree to which a job is potentially anomalous. This is due to the definition of anomaly score in Eq.~\eqref{eq:anomaly_score} and provides a set of k jobs that contribute the most to the instability of the workflow. Equipped with this information, the system administrator can further investigate the root cause of the anomaly. In order to enable this analysis, however, a threshold $\tau$ is a hyperparameter that needs to be tuned carefully.
We observed that the performance of our model is sensitive to the value of $\tau$, and the best performance is achieved when $\tau=0.5$.
Table~\ref{tab:topk} reports the top-k precision with different values of $k$ on the three workflows. We can see that a smaller $k$ enables  better precision and facilitates the detection of anomalous jobs with high confidence.
\begin{table}
  \centering
  \caption{Evaluation of top-K precision}
  \label{tab:topk}
  \begin{tabular}{c|cccc}
    \toprule
                         & $k=5$ & $k=10$ & $k=20$ & $k=\abs{\yvec_{\text{test}}}$ \\
    \midrule
    1000 Genome          & 0.8   & 0.7    & 0.55   & 0.43                          \\
    Montage              & 0.8   & 0.6    & 0.45   & 0.30                          \\
    Predict Future Sales & 0.6   & 0.6    & 0.55   & 0.35                          \\
    \bottomrule
  \end{tabular}
  \vspace{-10pt}
\end{table}

\subsubsection{Distributions on Latent Space}
\begin{figure}
  \centering
  \begin{subfigure}[b]{0.48\linewidth}
    \centering
    \includegraphics[width=\linewidth]{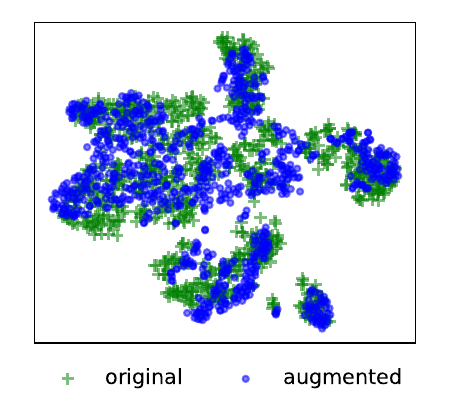}
    \caption{Normal distribution}
    \label{fig:aug_normal}
  \end{subfigure}
  \hfill
  \begin{subfigure}[b]{0.48\linewidth}
    \centering
    \includegraphics[width=\linewidth]{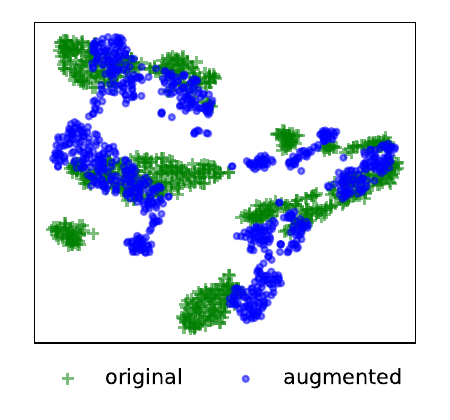}
    \caption{Gumbel distribution}
    \label{fig:aug_gumbel}
  \end{subfigure}
  \caption{t-SNE plot of the original and augmented data in the hidden space}
  \label{fig:augmented}
  \vspace{-10pt}
\end{figure}

Since part of our model utilize the augmented data for margin loss to facilitate the training, we would like to see how the augmented data is distributed in the hidden space.
We use t-SNE to visualize the hidden space of the original and augmented data.
Figure~\ref{fig:aug_normal} and Figure~\ref{fig:aug_gumbel} are the t-SNE plots of original and augmented data (a single minibatch sample from 1000 Genomes Project) in latent space with learnable Normal and Gumbel distributions, respectively.
The plots indicate that the augmented data is distributed around the original data, preserving the original data distribution while introducing variances in the hidden space.
This enhances the robustness and generalization of the training without explicit labeling information from original samples.
We observed that the Gumbel distribution is more suitable for margin loss, as shown by its better separability in Figure~\ref{fig:aug_gumbel} and verified by the performance in Table~\ref{tab:model_comp_unsupervised}.

\begin{figure}
  \centering
  \includegraphics[width=0.8\linewidth]{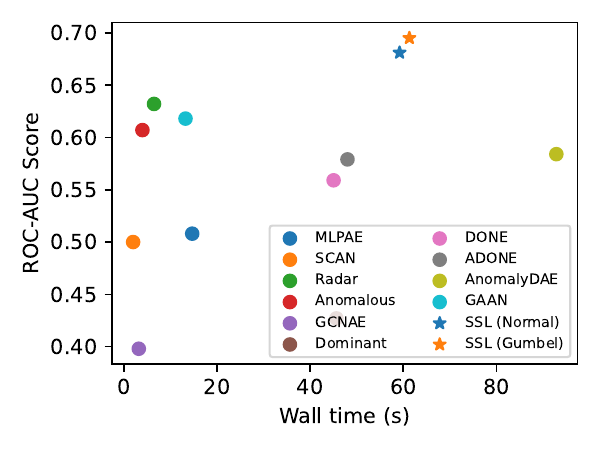}
  \caption{Time efficiency cost}
  \label{fig:time_comp}
  \vspace{-10pt}
\end{figure}

To compare the efficiency of our SSL model based on graph partition with other deep learning methods~(variants of graph neural network), we plot the wall time consumed for the 1000 Genomes workflow in Figure~\ref{fig:time_comp}. As shown in the figure, our SSL models, both with normal and Gumbel softmax trick in latent space, achieve better ROC-AUC scores while the training time is still comparable to that of other methods, even with the additional augmented data processing built into our model. This indicates that the mini-batch training with graph partition can speed up the training process without sacrificing performance, and this speed assists our SSL approach to achieve high precision without sacrificing scalability, which is in contrast with most anomaly detection approaches in the literature.

\section{Conclusion}
Detecting anomalies is a challenging problem in machine learning because of the absence of label information and the imbalanced nature of the data. To overcome these challenges in computational workflows, we have proposed a self-supervised learning approach using directed acyclic graphs to model the workflows. This approach uses augmented data to learn the representative latent space of unlabeled data. Our experiments with SSL, employing both normal and Gumbel softmax tricks in the hidden space, have produced promising results. These results have outperformed a set of benchmark methods that were applied to both tabular and graph data. We achieve better performance without sacrificing scalability.

Potential future work could focus on exploring online detection with temporal information from jobs and investigating the approach's performance on larger and more complex datasets. Moreover, the proposed approach could be extended to other domains beyond computational workflows, such as image and speech recognition, to further evaluate its effectiveness.

\section*{Acknowledgment}
This work is funded by the Department of Energy under the Integrated Computational and Data Infrastructure (ICDI) for Scientific Discovery, grant DE-SC0022328. This material is based upon work supported by the U.S. Department of Energy, Office of Science, under contract number DE-AC02-06CH11357.

\bibliographystyle{IEEEtran}
\bibliography{bibfile}

\end{document}